\documentclass[conference]{IEEEtran}
\IEEEoverridecommandlockouts
\usepackage{cite}
\usepackage{amsmath,amssymb,amsfonts}
\usepackage{algorithmic}
\usepackage{graphicx}
\usepackage{textcomp}
\usepackage{xcolor}
\usepackage{color}
\def\BibTeX{{\rm B\kern-.05em{\sc i\kern-.025em b}\kern-.08em
    T\kern-.1667em\lower.7ex\hbox{E}\kern-.125emX}}

\newcommand*{\myfont}{\fontfamily{<phv>}\selectfont}
\DeclareTextFontCommand{\textmyfont}{\myfont}

\usepackage{amsmath}

\DeclareMathOperator*{\argmin}{\arg\!\min}

\usepackage{tikzscale}
\usepackage{tikz}
\tikzset{font={\fontsize{8pt}{9}\selectfont}}
\usetikzlibrary{shapes}
\usetikzlibrary{petri}
\usetikzlibrary{arrows, decorations.markings}
\usetikzlibrary{decorations.pathreplacing,angles,quotes}
\usetikzlibrary{automata,positioning}

\usepackage{array}
\newcolumntype{L}[1]{>{\raggedright\let\newline\\\arraybackslash\hspace{0pt}}m{#1}}
\newcolumntype{C}[1]{>{\centering\let\newline\\\arraybackslash\hspace{0pt}}m{#1}}
\newcolumntype{R}[1]{>{\raggedleft\let\newline\\\arraybackslash\hspace{0pt}}m{#1}}

\definecolor{MyLightGreen}{RGB}{197,224,180}
\definecolor{MyPeach}{RGB}{248, 203, 173}

\tikzset{place/.style = {circle, draw=black!50, fill=blue!20, thick, minimum size=0.5cm},
	big_circle/.style = {circle, draw=black!50, thick, minimum size=4.5cm},
    my_blue_ellipse/.style = {ellipse, draw=blue!20, fill=blue!20, thick, minimum width=2cm, minimum height = 1.25cm},
    my_yellow_ellipse/.style = {ellipse, draw=yellow!20, fill=yellow!20, thick, minimum width=2cm, minimum height = 1.25cm},
    my_red_ellipse/.style = {ellipse, draw=red!20, fill=red!20, thick, minimum width=2cm, minimum height = 1.25cm},
    transition/.style = {rectangle, draw=black!50, fill=black!10, thick, minimum width=3cm, minimum height = 1cm},
    F_embedding/.style = {rectangle, draw=black!50, fill=blue!5, thick, minimum width=2cm, minimum height = 0.7cm},
    S_embedding/.style = {rectangle, draw=black!50, fill=green!5, thick, minimum width=2cm, minimum height = 1.5cm},
    Conv_embedding/.style = {rectangle, draw=black!50, fill=MyPeach, thick, minimum width=1.5cm, minimum height = 1cm},
    Attention/.style = {rectangle, draw=black!50, fill=black!5, thick, minimum width=3cm, minimum height = 0.75cm},
    head/.style = {rectangle, draw=black!50, thick},
    neuron/.style = {circle, draw=black!50, thick, fill=black!10, rounded corners=0.5mm},
    blueneuron/.style = {rectangle, draw=black!50, thick, fill=blue!10,rounded corners=0.5mm, minimum width=5.5cm, minimum height = 5.5cm},
    greenneuron/.style = {rectangle, draw=black!50, thick, fill=green!10,rounded corners=0.5mm, minimum width=5.5cm, minimum height = 5.5cm},
    output/.style = {rectangle, draw=black!50, thick, fill=black!10},
    pre/.style =    {<-, semithick},
    post/.style =   {->, semithick}
}

\tikzstyle{VecArrow} = [thick, decoration={markings,mark=at position
   1 with {\arrow[semithick]{open triangle 60}}},
   double distance=2pt, shorten >= 5.5pt,
   preaction = {decorate},
   postaction = {draw, line width=1.6pt, white, shorten >= 4.5pt}]
\tikzstyle{innerWhite} = [semithick, white,line width=1.4pt, shorten >= 4.5pt]

\usetikzlibrary{decorations.pathreplacing,angles,quotes}

\begin{document}

\title{Learning to Support: Exploiting Structure Information in Support Sets for One-Shot Learning\\
}

 \author{\IEEEauthorblockN{Jinchao Liu}
 \IEEEauthorblockA{
 Canterbury, UK \\
 liujinchao2000@gmail.com}
 \and
 \IEEEauthorblockN{Stuart J. Gibson}
 \IEEEauthorblockA{\textit{School of Physical Sciences} \\
 \textit{University of Kent}\\
 Canterbury, UK \\
 S.J.Gibson@kent.ac.uk}
 \and
 \IEEEauthorblockN{Margarita Osadchy}
 \IEEEauthorblockA{
 \textit{Department of Computer Science} \\
 \textit{University of Haifa}\\
 Haifa, Israel \\
 rita@cs.haifa.ac.il}
 }

\maketitle

\begin{abstract}
Deep Learning shows very good performance when trained on large labeled data sets. The problem of training a deep net on a few or one sample per class requires a different learning approach which can generalize to unseen classes using only a few representatives of these classes. This problem has previously been approached by meta-learning. Here we propose a novel meta-learner which shows state-of-the-art performance on common benchmarks for one/few shot classification. Our model features three novel components: First is a feed-forward embedding that takes random class support samples  (after a customary CNN embedding) and transfers them to a better class representation in terms of a classification problem. Second is a novel attention mechanism, inspired by competitive learning, which causes class representatives to compete with each other to become a temporary class prototype with respect to the query point. This mechanism allows switching between representatives depending on the position of the query point. Once a prototype is chosen for each class, the predicated label is computed using a simple attention mechanism over prototypes of all considered classes. The third feature is the ability of our meta-learner to incorporate deeper CNN embedding, enabling larger capacity. Finally, to ease the training procedure and reduce overfitting, we  averages the top $t$ models (evaluated on the validation) over the optimization trajectory. We show that this approach can be viewed as an approximation to an ensemble, which saves the factor of $t$ in training and test times and the factor of of $t$ in the storage of the final model.
\end{abstract}


\section{Introduction}
Deep Learning (DL) has shown very impressive results in various domains, such as image/video, speech, and text, showing human level performance in some of these.
However, there are still a number of open problems that limit the use of DL. One of these problems is the requirement of very large labeled training sets. The performance of DL drops rapidly when the training set size decreases. The requirement of large labeled sets is expensive in terms of data collection and training time. In practice, many learning problems require rapid inference from small amounts of data. In particular, practical systems should be able to recognize a new category from a handful of training images. Moreover, it is easy for a human to recognize novel images of an unseen object after seeing only a single representative of that object. Motivated by this ability, recent work started to explore various directions in machine one/few-shot learning (e.g,~\cite{Koch2015SiameseNN,MatchingNet,PrototypeNet,Snail,FinnAL17,RelationNet}).

As deep networks can extract very good features, but require large training sets, the research has focused on designing auxiliary learning tasks that utilize the available training data smartly. For example, for a multi-class problem with many classes, but only few samples per class, one can consider an auxiliary task of identifying pairs of inputs as belonging to the same or different classes. Hence the task is formulated as a binary classification problem with sufficient data available for representation learning \cite{Koch2015SiameseNN}. A somewhat different auxiliary task, used for one/few-shot learning and termed meta-learning, learns a strategy for generalizing to an unseen task from a similar task distribution (e.g.,~\cite{MatchingNet,PrototypeNet,Snail,FinnAL17}). Here instead of learning the distribution of samples, it learns the distributions of tasks, which explains the name meta-learner.

The meta-learning architectures can be categorized into three methods~\cite{NIPS_Symposium}: 1) optimization-based,  2) model-based, and 3) metric-based.
The strategy in optimization-based meta-learning (e.g.,~\cite{Andrychowicz,LiM16b,FinnAL17,Sachin2017}) is to optimize a given network such that it can be effectively fine-tuned on a small data set within a few gradient-descent updates. The model-base methods (e.g,~\cite{santoro16,MunkhdalaiY17})  try to relate unseen tasks to those that they saw in the past and memorized in the internal states of an RNN or other network with memory. The metric-based methods (e.g.,~\cite{Koch2015SiameseNN,MatchingNet,PrototypeNet,RelationNet}) learn a comparator, which can take a form of a weighted nearest-neighbor with a fixed or learned metric, or a more complex non-linear learnable similarity function.

Most of the methods in one/few-shot learning embed the input samples using deep embedding. As deeper models have been shown to perform better in standard learning problems~\cite{VGG,Inception,ResNet}, it seems that integrating deeper embeddings in the  meta-learning framework should also be possible. The results in~\cite{Snail} clearly showed the merit of using deep ResNet~\cite{ResNet} withing the meta-learning framework. Unfortunately, most of  the previous meta-learners could only afford a relatively shallow embedding.

Here we propose a new method for one/few-shot learning that falls into the category of metric-based methods. The following discussion provides context for our approach in relation to previous work. Most of metric-based methods use a \textit{support set} of labeled samples to learn predictions for labels of unseen samples, the \textit{query set}. To extract good features, all samples are embedded using deep (usually convolutional) embedding and then the embedded query is compared to the embedded support set. The simplest way of doing this is by using a weighted nearest neighbor classifier with a fixed metric~\cite{MatchingNet}. This simple approach can be improved using a learnable comparator, which could take the form of a linear classifier~\cite{PrototypeNet} or metric learning~\cite{Koch2015SiameseNN}. Alternatively, it could be a more complex non-linear comparator as in e.g.,~\cite{RelationNet,MatchingNet,Snail}.

\def\layersep{2.5cm}
\begin{figure*}
\centering
\includegraphics[width=0.85\textwidth]{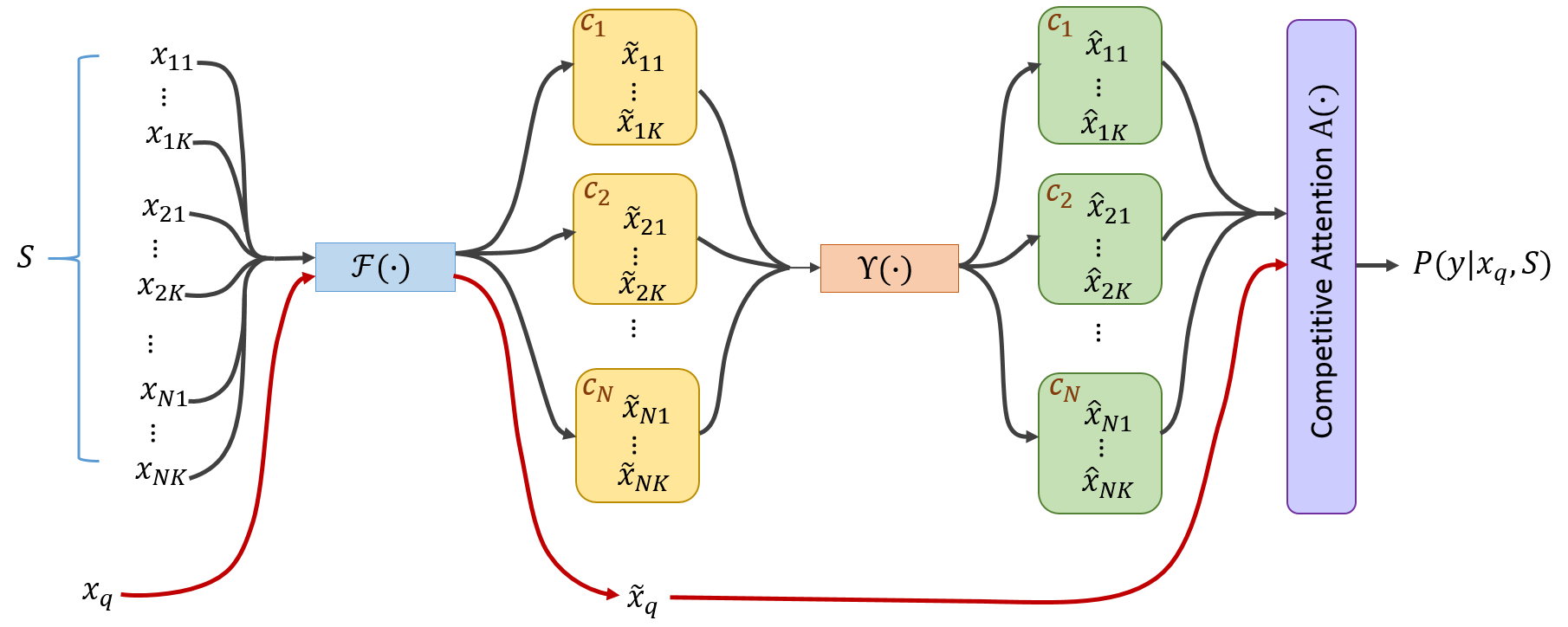}
\caption{Diagram of the Class Support Network. $\{(x_{i,j}, y_{i}), i=1,\cdots,N, j=1,\cdots,K)\}$ forms a support set; $x_{q}$ denote a query point; $\mathcal{F}$ denotes the CNN embedding, which applies to all the samples individually. $\Upsilon$ denotes the proposed class support set embedding, which inputs subsets of the support set belonging to the same class (denoted as $\{(\tilde{x}_{i,j}, y_{i}), i=1,\cdots,N, j=1,\cdots,K)\}$) and outputs the new class support (denoted as $\{(\hat{x}_{i,j}, y_{i}), i=1,\cdots,N, j=1,\cdots,K)\}$). $\mathcal{A}$ denotes the new competitive attention mechanism on the support set that outputs label probability.}
\label{Fig:LSNDiagram}
\end{figure*}

\def\layersep{2.5cm}
\begin{figure}
\centering
\includegraphics[width=0.3\textwidth]{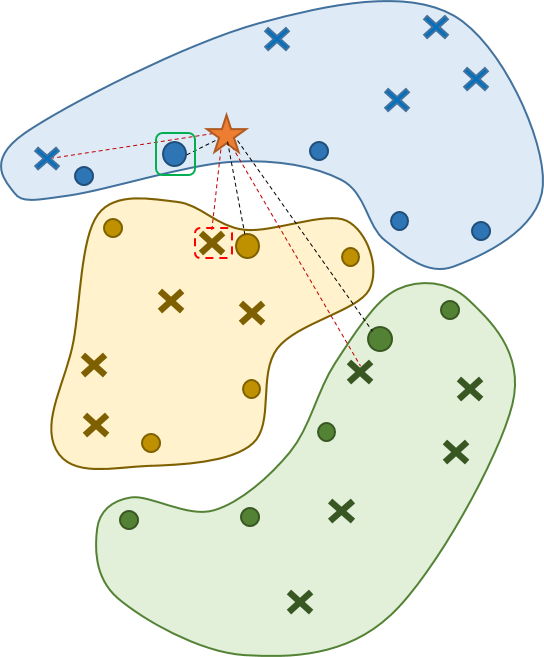}
\caption{Illustrative example of class support embedding. x's denote a random support set, color codded by class. A query point, shown as a star, is classified incorrectly using a weighted K-nearest neighbor (here K=1 for simplicity). The distances are shown in red dotted lines. The proposed class support embedding approximates the boundary of the classes (using the meta-learning) and outputs a more discriminative set of class representatives, shown in circles. Using  the new class support, any internal point in all three classes will have the closest representative from its own class. Namely, the distances shown as black dotted lines from the star point to the closest representatives from the updated set provide a good measure for classification. Larger circles represent the closest neighbor of the query point in each class. We can see, that using the closest point as a class prototype is better than averaging all class support points.}
\label{Fig:Class_support_Illustration}
\end{figure}

\def\layersep{2.5cm}
\begin{figure*}
\centering
\includegraphics[width=0.75\textwidth]{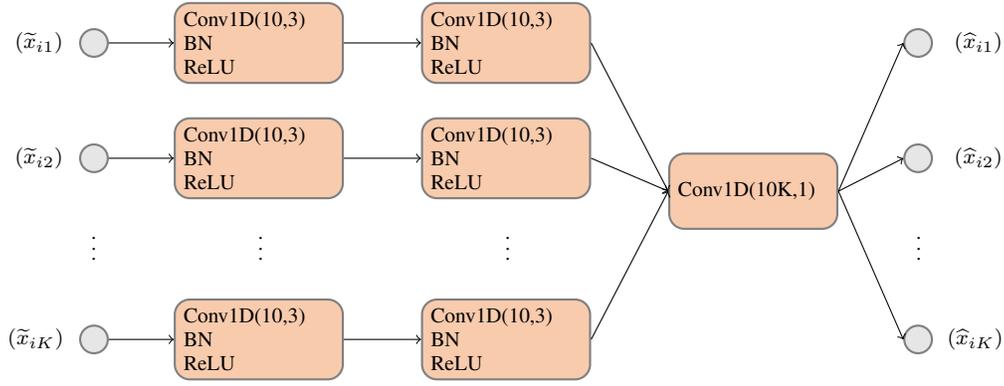}
\caption{Class support embedding architecture. All convolutional kernels in the network are one dimensional. We use Conv1D(m,n) to denote m one-dimensional convolution filters with kernel size n. BN stands for Batch Normalization.}
\label{Fig:SSEDiagram}
\end{figure*}

Instead of comparing a query to all members in the support set, it was suggested in~\cite{PrototypeNet} to compute a class prototype (by simple averaging) of the support elements belonging to the same class and use these prototypes for classifying the query. One can view this as a very simple support set embedding. The mean of class support (after embedding) was also used in~\cite{NeuralStat}  to provide a prior to a variational encoder which parameterized a diagonal Gaussian, serving as an approximated class posterior. The classification of a query was done by choosing the class which minimizes the Kullback Leibler divergence between the approximated posterior of the query and the approximated posterior of each class. A more complex embedding of the support set was proposed in~\cite{MatchingNet,Snail}. In ~\cite{MatchingNet} a bidirectional LSTM, named the full context embedding (FCE), was learned to model the dependences in the support set. The model, in~\cite{Snail} combined temporal convolutions, which aggregate contextual information from past experience, with causal attention, which makes the information more specific.

As can be observed from the results reported in these works, support set context embedding has obvious benefits. While the type of the embedding varies among methods,  all of them have their advantages and shortcomings. For example, representing classes with prototypes, obtained from the support set, is beneficial over plain comparison to the entire support set~\cite{PrototypeNet,NeuralStat}. Integrating context embedding of the support set has a clear advantage over simpler approaches that apply a learned metric~\cite{Koch2015SiameseNN} or weighted K-nearest neighbor~\cite{MatchingNet} on the initial embeddings. However, making the context embedding of the support set too complex, e.g.,  bidirectional LSTM on the entire support set~\cite{MatchingNet}, has an obvious downside. As the number of classes and shots increases, the model is required to learn longer and more complex dependences, which negatively affects both generalization and efficiency.

Here, we use these observations to build an architecture that learns a simpler model for metric-based meta-learning.  This model includes a feed-forward context embedding of the \emph{class support} (shots of the same class), as opposed to the context embedding of the entire support set. This new class support is used to classify the query set using a novel attention mechanism that combines competitive learning~\cite{Comp_leanr} with attention based classification~\cite{MatchingNet}. We learn the same class support embedding for all classes (tasks) in the training set, thus all classes share the knowledge of combining  random support elements of a class to produce better class representatives. This shared knowledge can be regarded as meta-knowledge. Our experiments show a clear advantage of class-level support embedding over previous approaches.

We also provide a statistical view of the proposed class-level embedding. A support set is sampled at random, thus the elements of the support set, regardless of the name, do not provide the optimal (or even good) support of the classes. 
Our proposed class support embedding is aimed at correcting this problem. Specifically, during episode training, the class-level embedding sees many sets of points from different classes, which allows it to learn the distribution of the entire space. Thus, given an approximate location of the classes via a random support set, the network produces a corrected set of class representatives, which takes into account the knowledge of the distribution that the network learned from many samples of support sets. The experiments show that our novel meta-learning approach enables a deep embedding and improves capacity without overfitting.

Finally, we observed that choosing the stopping criterion in training (the common practice in training meta-learning methods) is difficult. To address this problem and reduce overfitting, we suggest applying an ensemble of models, in which the models' diversity comes naturally from the episode training. This allows models from the same optimization trajectory to be used to form an ensemble and avoid the increase in training time of learning different models. Specifically, we propose choosing the best $t$ models from the same optimization trajectory using the validation set. We also show that this ensemble can be approximated (with very small accuracy loss) by averaging the parameters of these best models, thereby making a time and storage saving of factor $t$.

\section{Class Support Networks}
\subsection{Model}
Let $S=\{(x_{i,j}, y_{i})| i=1,\cdots,N; j=1,\cdots,K)\}$ denote a support set and $S_{i}\in S$, $i=1,\cdots,N$ denote a subset of the support set belonging to class $i$. Let $\{x_{q}, y_{q}\}$ denote a query point and its label.
Formally, our model can be described as
\begin{align}
    &P(y_{q} | x_{q}, S) = \nonumber \\
    &\mathcal{A} \left[ d\Big(\Upsilon_{\phi} (\mathcal{F_{\theta}}(S_{1})), \cdots, \Upsilon(\mathcal{F_{\theta}}(S_{N})), \mathcal{F_{\theta}}(x_{q}) \Big)\right]
    \label{Eq:Model}
\end{align}
where $\mathcal{F}$ denotes a deep embedding which is applied to all samples individually, $\Upsilon$ denotes the proposed \emph{class support embedding}, $d$ is a metric, e.g. $L_{2}$ distance, and $\mathcal{A}$ denotes the novel attention mechanism on the support set.
A diagram of the class support network is depicted in Fig. \ref{Fig:LSNDiagram} and the descriptions of the class support embedding and the novel competitive attention mechanism are presented in Sections~\ref{Sec:class_level_emd} and~\ref{sec:attention} respectively.

The training objective is to minimize the negative log-likelihood in Eq.~\ref{Eq:Model}:
\begin{align}
    \Theta = \argmin_{\Theta}   \sum_{(x,y),S} \log  P(y_{q} | x_{q}, S)
    \label{Eq:ModelLoss}
\end{align}
where $\Theta=[\theta,\phi]$ are the trainable parameters of the embeddings $\mathcal{F_{\theta}}$ and $\Upsilon_{\phi}$. In our implementation, both $\mathcal{F_{\theta}}$ and $\Upsilon_{\phi}$ are realized as convolutional neural networks (CNN) and the entire model in Eq.~\ref{Eq:Model} is trained end-to-end using stochastic gradient descent (SGD).

\subsection{Class Support Embedding}\label{Sec:class_level_emd}
The support set $S$ includes random samples of $N$ classes.  These samples do not provide the optimal representation of the corresponding classes or the boundaries between them. To address this problem, we propose learning a non-linear function that inputs elements of the support set from the same class and outputs a new class support, which is better suited for the given classification task. Since the assumption is that all tasks in the distribution  are related, we can represent this function as a neural network and learn its parameters using episode training. Fig.~\ref{Fig:Class_support_Illustration} schematically illustrates this idea.

Formally, the class support embedding is defined as
$$\Upsilon:\tilde{S}_i \rightarrow \hat{S}_i, \;\;\; i=1,...,N$$
where $\tilde{S}_i=\mathcal{F_{\theta}}(S_i)$ and  $\hat{S}_i$ is the new support set for class $i$. We note that the size of $\hat{S}_i$ could differ from the size of $\tilde{S}_i$, but in this work, we set both sets to be of size $K$ (the number of shots).

Class support embedding can be realized by any differentiable model which maps a set to another set, e.g. MLP, CNN, seq2seq. CNN has been shown to train better and resist overfitting, therefore we propose to realize the class support embedding by CNN. The architecture of the class support embedding is shown in Fig.~\ref{Fig:SSEDiagram}.
It consists of three convolutional layers. The first two layers have identical structure comprising  a convolutional block, batch normalization, and non-linear activation ReLU. As the input tensors are one-dimensional representations of support points, the convolutional filters are one dimensional of size $1 \times 3$. To reduce the total amount of learnable parameters and prevent overfitting, the first two layers embed each input $\tilde{x}_{i,j}$ individually (there are no connections between the internal representations of different inputs).
The third block includes a one dimensional convolutional layer of kernel size 1. This block integrates information from the internal representations of all inputs.

\subsection{Competitive Attention}\label{sec:attention}
While one can consider using attention mechanisms suggested in previous work, e.g.,~\cite{MatchingNet}, we argue that the use of class support embedding requires a special attention mechanism.

We rely on the following intuition as motivation for the attention mechanism that we propose here and we use the example depicted in Fig.~\ref{Fig:LSNDiagram} to illustrate it. We suggest that the class support embedding outputs points that better represent the boundaries of the class.  This, however, could result in the situation where support points belonging to the same class are pushed to the boundary on the opposite sides of the class. In that case, it makes more sense to select the support point closest to a query point to represent the class within the nearest neighbor classification, than using a weighted distance from the query to all support points in the class (as in~\cite{MatchingNet}). Following this intuition, we want the class support points to compete with each another to represent the class in terms of their distances to the query point. This is a typical scenario for competitive learning\cite{Rumelhart1986}.

Competitive learning is a variant of Hebbian learning which takes roots in neuroscience\cite{hebb-organization-of-behavior-1949}, but recently has shown potential in deep neural networks\cite{PlausibleHebbianLearning, DeepHebbian}. The simplest form of the competitive learning performs rounds of training in which neurons compete for the right to be activated. In each round the model is restricted to have only one activated neuron, while the rest must remain silent.

Inspired by the connection to competitive learning, we named the proposed attention mechanism, \emph{competitive attention}.
Given a new set of class support points $\{ \widehat{x}_{11}, \cdots, \widehat{x}_{NK} \}$ (produced by the class support embedding $\Upsilon_{\phi}$), and a query point $\widetilde{x}_{q}$ (where $\widetilde{x}_{q}=\mathcal{F_{\theta}}(x_q)$), the competitive attention mechanism is defined as follows:
 \begin{align}
 \label{Eq:AttentionKernel_1}
 	a(\widehat{x}_{i, J^{*}_{i}}, \widetilde{x}_{q}) = \frac{\exp{\|\widehat{x}_{i, J^{*}_{i}} - \widetilde{x}_{q}\|} } {\sum_{i} \exp{\| \widehat{x}_{i, J^{*}_{i}}- \widetilde{x}_{q}\|}}
 \end{align}
 where
 \begin{align}
  \label{Eq:AttentionKernel_2}
 	J^{*}_{i} = \argmin_{j} \big \{ \| \widehat{x}_{ij} - \widetilde{x}_{q}  \|, j=1,\cdots,K \big \}
 \end{align}
Support points belonging to the same class first compete with one another to represent the class as in Eq.~\ref{Eq:AttentionKernel_2}. The indexes of the winners form a set $J^*=\{J^{*}_{i}|i=1,\cdots,N\}$. Then, the winning points of $N$ classes produce a weight distribution collaboratively as in Eq.~\ref{Eq:AttentionKernel_1}. Finally, the predicted label is obtained as:
\begin{align}
 	\hat{y}_q=\sum_{t\in J^*}a(\widehat{x}_{i,t},\widetilde{x}_{q})y_t
 \end{align}

\begin{figure}
\centering
\includegraphics[width=\columnwidth]{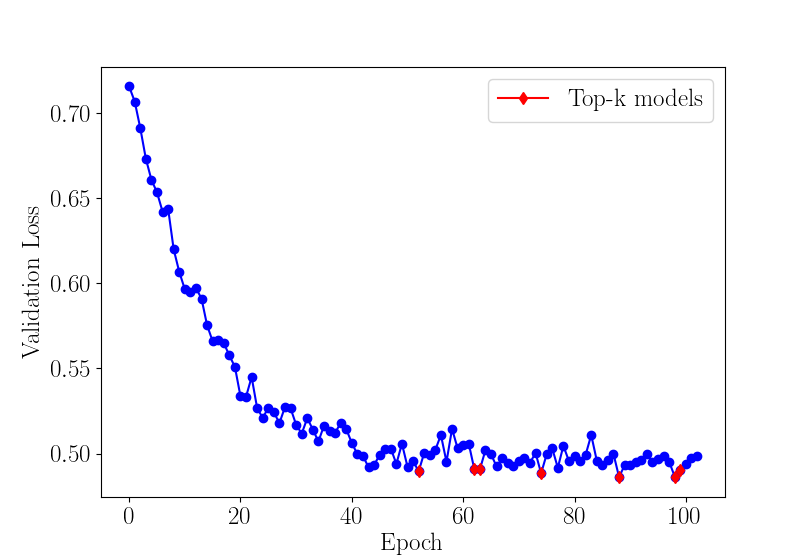}
\caption{Illustration of the proposed ensemble method which comprises the best $t$ models on the validation set (marked in red stars). Averaging these models achieves a better result than a single model, obtained by early stopping. }
\label{Fig:MPA_loss}
\end{figure}

\subsection{Optimizing Meta-Learners with Ensemble Learning}
Most of the previous meta learning methods followed a standard training protocol of decreasing learning rates and applying early stopping using the validation set. However, finding a good stopping criterion is quite involved as the validation error osculates and hardly decreases as shown in Figure~\ref{Fig:MPA_loss}. This happens because different models are better on different sub-sets of episodes. Ideally, we would like to learn a model that can generalize to all episodes, but it doesn't happen in practice on the existing benchmarks. This may be remedied by using ensemble methods. However, using ensembles  will dramatically increase the training time and test time as it requires training and running multiple models. It was suggested in~\cite{SnapshotHuang} that multiple models can be obtained from a single training cycle by causing the optimization to visit several local minima (this was done by alternating the learning rates). The assumption used in~\cite{SnapshotHuang} was that models corresponding to different minima are diverse enough to comprise an effective ensemble.

We follow the idea of obtaining different models from the same optimization path. However, we believe that in episode training models are diverse because they see different episodes. This introduces natural model diversity and removes the need to insert diversity artificially (as in e.g,~\cite{SnapshotHuang}).  To see that this is the case, consider 5-way classification, trained on 100 classes with 5 samples per class. The total number of all possible episodes amounts to  $3\times10^{13}$. A typical training process goes through 300K randomly sampled episodes. This is only a small fraction ($10^{-8}$) of the total amount of episodes. Thus different models on the same optimization path are different due to additional episodes that they saw.

Unfortunately, it seems that training on more episodes from the same set of classes causes overfitting (as they see same classes). Instead, we propose to choose $t$ best models on the validation set (from the same optimization trajectory) and use them to form the ensemble.

This simple approach creates an effective ensemble with no additional training cost. 
We show in Appendix~\ref{AppendA} that under the model linearity assumption, averaging model predictions is the same as averaging models' parameters. Our network is not a linear function, thus the linearity assumption may not hold. However, we suggest that the models' average can be used as a crude approximation of the ensemble. Formally, we select $t$ best points along the trajectory and average their parameters equally to produce a single solution: $\widehat{\Theta} = \frac{1}{t}\sum_{i=1}^{t}\Theta(B(i))$, where $B(i)$ is the index of the $i^{th}$ best point/model. We refer to this method as \textit{Approximate Ensemble of Meta-Learners} (AEML). Our experiments reported in~\ref{AppendB} demonstrate that such an approximation is possible and reduces the test time and the storage by a factor of $t$ (the number of models in the ensemble). Moreover, we observed in our experiments that AEML effectively avoided the need for careful design of the early stopping conditions and made it easier to train deeper models.

\begin{figure}
\centering
\includegraphics[width=0.6\columnwidth]{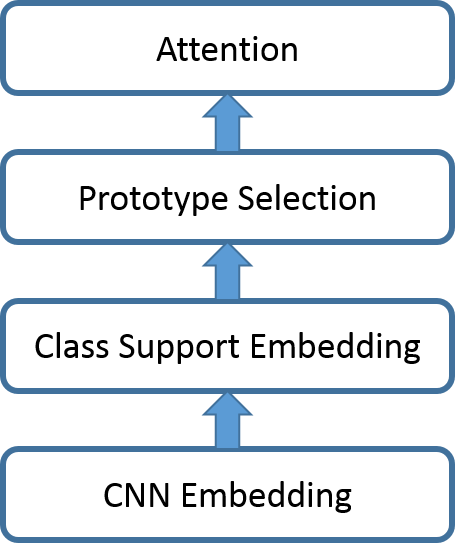}
\caption{Block Diagram of the Class Support Network }
\label{Fig:CSN_block}
\end{figure}
\section{Relation to Previous Work}
In this section we discuss the relation of our work to methods in meta-learning and  to Nearest Neighbor-based metric learning.
\subsection{Meta-Learning}
We can illustrate our meta-learner by a block diagram in Fig.~\ref{Fig:CSN_block}. It consists of a CNN embedding block, class support embedding block, prototype selection, and attention block (the later two form the competitive attention mechanism). The CNN embedding has been used in most of the recent meta-learners. The class support block is novel. The most related work to this block is Matching Networks with FCE~\cite{MatchingNet}. In our work the embedding is done at the class level, while FCE operates on the entire support set. Our class embedding is implemented as a simple feed-forward network, which is very easy to train and its complexity does not depend on the number of classes. In contrast, FCE is implemented as a bidirectional LSTM and its complexity grows with $N$.

The prototype selection block  is related to Prototypical Networks~\cite{PrototypeNet} in which the prototype is computed as an average of class representatives. This implicitly assumes that the distribution of classes is Gaussian. In our work we choose the prototype dynamically with respect to the query point using the concept of competitive learning. This way we avoid making assumptions about the distribution of the class.

The attention block is very similar to the one introduced in Matching Nets~\cite{MatchingNet}, with the difference that it operates on the reduced support set comprising a single prototype per class, while the Attention Kernel in~\cite{MatchingNet} attends the entire support set.

After excluding the class support embedding block from our network, it becomes equivalent to the Matching and Prototype Nets for one-shot learning. Our experiments show that our method (with all blocks) outperforms these two models for one-shot learning, which verifies the importance of the proposed class support embedding.

\subsection{Nearest Neighbor-Based Metric Learning}
We note that our method is related to Large Margin Nearest Neighbor (LMNN) \cite{Weinberger2009} which is a very successful method for metric learning. In contrast to our method that learns non-linear transformation of class support points, LMNN learns a Mahalanobis metric, which is linear. The similarity is in the LMNN training objective, which is explicitly defined to reposition the data points thereby facilitating k-nearest neighbor classification. However, the meta-knowledge that LMNN learns is limited to transforming all data points in a fixed and static manner. Our method also learns to reposition the support points to achieve nearest neighbor classification, but the non-linear mapping in collaboration with competitive prototype selection allows a better adaptation to the given classification task.

\renewcommand{\arraystretch}{1.25}
\begin{table*}[!htp]
\caption{One/Few Shot classification accuracies on Omniglot. All
accuracy results are averaged over 2000 test episodes and are reported
with 95\% confidence intervals. CNN stands for the CNN embedding used in the compared methods.}
\begin{center}
\begin{tabular}{|c|C{1.5cm}|c|c|c|c|}
\hline
\textbf{Model}& \textbf{CNN} & \multicolumn{2}{|c|}{\textbf{5-way Accuracy}} & \multicolumn{2}{|c|}{\textbf{20-way Accuracy}}\\
\cline{3-6}
& \textbf{Embedding} & \textbf{1-shot} &  \textbf{5-shot} & \textbf{1-shot} & \textbf{5-shot} \\ \hline
\textbf{SIAMESE NETS}\cite{Koch2015SiameseNN} & 4 & 96.7\% & 98.4\% & 88.0\% &  96.5\%
\\
\textbf{MATCHING NETS}\cite{MatchingNet} & 4 & 98.1\% & 98.9\% & 93.8\% &  98.5\%
\\
\textbf{NEURAL STATISTICIAN}\cite{NeuralStat} & 4 & 98.1\% & 99.5\% & 93.2\% &  98.1\%
\\
\textbf{PROTOTYPICAL NETS}\cite{PrototypeNet} & 4 & 98.8\% & 99.7\% & 96.0\% &  98.9\%
\\
\textbf{MAML}\cite{FinnAL17} & 4 & 98.7$\pm$0.4\% & \textbf{99.9$\pm$0.1\%} & 95.8$\pm$0.3\% &  98.9$\pm$0.2\%
\\
\textbf{RALATION NETS}\footnotemark \cite{RelationNet} & 4 & 99.6$\pm$0.2\% & 99.8$\pm$0.1\% & 97.6$\pm$0.2\% &  99.1$\pm$0.1\% \\
\textbf{SNAIL}\cite{Snail} & 4 & 99.07$\pm$0.16\% & \textbf{99.78$\pm$0.09\%} & 97.64$\pm$0.30\% &  \textbf{99.36$\pm$0.18\%}\\  \hline
\textbf{CLASS SUPPORT NETS(OURS)} & 4 & \textbf{99.24$\pm$0.14\%} & \textbf{99.75$\pm$0.15\%}  & \textbf{97.79$\pm$0.06\%} & \textbf{99.27$\pm$0.15\%} \\
\textbf{CLASS SUPPORT NETS(OURS)} & 6 & 99.37$\pm$0.09\% & 99.80$\pm$0.03\% & 98.58$\pm$0.07\% & 99.45$\pm$0.04\% \\ \hline
\end{tabular}
\label{Tab:Omniglot}
\end{center}
\end{table*}

\section{Experiments}
We follow the standard practices of training a meta-learner for one/few shot learning using \textit{episodes}. Each episode consists of a support set and a set of query points and represents a classification task. Several episodes may form a mini-batch for SGD.

Similarly to previous works, we generate training episodes that match \textit{way} and \textit{shot} of the classification tasks in test time. For instance, for $N$-way $K$- shot classification, we generate $N$-way $K$-shot episodes for training and validation.
\subsection{Results}
\subsubsection{Omniglot}
Omniglot\cite{Lake_oneshot} consists of 50 different alphabets and 1623 characters in total. Each character has 20 samples which are hand drawn by 20 different people. Following the procedure of \cite{MatchingNet}, which has been employed by most existing works, we augment the characters (classes) by rotating the images 90, 180, 270 degrees. We use 1200 characters with the rotation augmentation for training and the remaining 423 characters with the rotation augmentation for testing. All images are resized to $28 \times 28$.

We use the CNN embedding proposed in \cite{MatchingNet} and adopted in most previous works. This embedding consists of four convolutional blocks, each of which comprises  64 convolutional filters of the size $3 \times 3$, followed by batch normalization, ReLU activation function, and a $2 \times 2$ max-pooling.

We validated the proposed Class Support Network in $N$-way $K$-shot classification tasks with $N=5,20$ and  $K=1,5$. Following our training protocol, we generated $N$-way $K$-shot episodes respectively. Each episode contained 10 query points per class. We trained our models end to end from scratch using the Adam optimizer with initial learning rate of $10^{-3}$ which decays by half for every 50K episodes.

Results are shown in Table \ref{Tab:Omniglot}. For the four classification tasks, the proposed method has achieved comparable results to state-of-the-art methods. We can observe that the results of recent methods on this data set have reached a very high accuracy, which makes this set a poor benchmark for further research in one/few-shot learning. The only setting which is not yet saturated is 20-way 1-shot classification. In this task our network with a standard 4-layer embedding achieved 97.79\%. However, by increasing the embedding to six convolutional layers and keeping all other settings/parameters unchanged, our network reaches 98.58\% which is significantly better than the state-of-the-art.

\subsubsection{\textit{mini}ImageNet}

\renewcommand{\arraystretch}{1.25}
\begin{table*}[htbp]
\caption{One/Few Shot classification accuracies on miniImagenet. All
accuracy results are averaged over 2000 test episodes and are reported
with 95\% confidence intervals. No fine tune has been used for the compared methods, except for MAML.}
\begin{center}
\begin{tabular}{|c|C{1.5cm}|C{1.5cm}|c|c|}
\hline
\textbf{Model}& \textbf{CNN} & \textbf{Fine Tune}& \multicolumn{2}{|c|}{\textbf{5-way Accuracy}} \\
\cline{4-5}
&  \textbf{Embedding} & \textbf{} & \textbf{1-shot} & \textbf{5-shot} \\
\hline
\textbf{META-LEARN LSTM}\cite{Sachin2017} & 4 & N& 43.44$\pm$0.77\% & 60.60$\pm$0.71\% \\
\textbf{MATCHING NETS\cite{MatchingNet}} & 4 & N& 43.56$\pm$0.84\% & 55.31$\pm$0.73\%  \\
\textbf{MAML}\cite{FinnAL17} & 4 & Y & 48.70$\pm$1.84\% & 63.11$\pm$0.92\% \\
\textbf{META NETS} & 5 & N& 49.21$\pm$0.96\% & - \\
\textbf{PROTOTYPICAL NETS\cite{PrototypeNet}} & 4 & N& 49.42$\pm$0.78\% & 68.20$\pm$0.66\% \\
\textbf{RELATION NETS}\cite{RelationNet} & 4 & N& 50.44$\pm$0.82\% & 65.32$\pm$0.70\% \\ \hline
\textbf{CLASS SUPPORT NETS (NO MPA, OURS)} & 4 & N & \textbf{52.50$\pm$0.45\%} & \textbf{68.46$\pm$0.37\%}\\
\textbf{CLASS SUPPORT NETS (OURS)} & 4 & N &  \textbf{52.56$\pm$0.45\%} & \textbf{69.05$\pm$0.36\%}\\ \hline \hline
\textbf{SNAIL}\cite{Snail} & 13 & N & 55.71$\pm$0.99\% & 68.88$\pm$0.92\% \\
\textbf{CLASS SUPPORT NETS (OURS)} & 9 & N &  \textbf{56.32$\pm$0.47\%} & \textbf{71.94$\pm$0.37\%}\\
\hline
\end{tabular}
\label{Tab:MiniImageNet}
\end{center}
\end{table*}

\renewcommand{\arraystretch}{1.25}
\begin{table*}[htbp]
\caption{Ablation analysis on miniImagenet. All
accuracy results are averaged over 2000 test episodes and are reported
with 95\% confidence intervals. Four layers CNN embedding were used. CS-NETS stands for the proposed class support networks. SE stands for support set embedding. No fine tune has been used.}
\begin{center}
\begin{tabular}{|c|c|c|c|c|c|c|c|}
\hline
\textbf{Model}& \textbf{CNN} & \multicolumn{3}{|c|}{\textbf{5-way Accuracy, without AEML}} & \multicolumn{3}{|c|}{\textbf{5-way Accuracy, with AEML}}\\
\cline{3-8}
& \textbf{Embedding} & \textbf{1-shot} & \textbf{2-shot} & \textbf{5-shot} & \textbf{1-shot} & \textbf{2-shot} & \textbf{5-shot} \\
\hline
\textbf{CS-NETS without SE} & 4 & 48.44$\pm$0.45\% & 57.64$\pm$0.42\% & 67.65$\pm$0.38\% & 48.85$\pm$0.43\% & 57.76$\pm$0.43\%  & 67.68$\pm$0.36\%
\\ \hline
\textbf{CS-NETS} & 4 & 52.50$\pm$0.45\%  & 59.86$\pm$0.42\% &68.46$\pm$0.37\%  & 52.56$\pm$0.45\% & 60.38$\pm$0.42\% & 69.05$\pm$0.36\%\\ \hline
\end{tabular}
\label{Tab:Ablation}
\end{center}
\end{table*}

\footnotetext{Note that according to their implementation, \cite{RelationNet} followed a different train/test evaluation protocol and their Omniglot results therefore may not be directly comparable to other methods in the table \ref{Tab:Omniglot}. For completeness, we include their results.}

The \textit{mini}ImageNet dataset is a subset of \textit{ILSVRC-12} dataset\cite{ILSVRC15} which was first proposed by Vinyals et al. \cite{MatchingNet}. It consists of 100 classes each of which contains 600 natural images of size $84 \times 84$\footnote{Note that much deeper embedding, hence better performance, may be possible if higher resolution images e.g. $200 \times 200$ are used. However, to ensure a fair comparison to prior work, the down-sampled image resolution should be kept as $84 \times 84$.}. We follow the splits of the \textit{mini}ImageNet, introduced by Ravi and Larochelle \cite{Sachin2017} and adopted in most previous work on few/one-shot learning.
The split uses 64 classes for training, 16 for validation and 20 for testing. We report the results in Table \ref{Tab:MiniImageNet}.

It was argued in~\cite{Snail}, that a deeper CNN embedding does not automatically translate to a better performance as the embedding comes in combination with the meta-learner. The meta-learner either exploits the capacity of the embedding or does not (depending on the method). The latter case leads to heavy overfitting. Thus recent work that used deeper CNN embedding~\cite{Snail}, reported the results separately for the standard 4-layer embedding and for the deeper one.

Following these practices, we performed two sets of experiments: with the 4-layer convolutional embedding (same as in the  Omniglot experiment) and with the deeper 9-layer convolutional embedding. The 9-layer embedding is a variant of LeNet\cite{Lecun98gradientbasedlearning} with a double-pyramid shape. It consists of four blocks, each followed by a $2 \times 2$ max-pooling. Each block includes two identical $3 \times 3$ convolutional layers with batch normalization. The number of filters in consecutive blocks of the embedding are 64, 128, 256 and 512. The last convolutional layer uses 2000 $1\times1$ filters, followed by an average pooling to control the size of the CNN features.

As shown in Table \ref{Tab:MiniImageNet}, our method with 4-layer embedding significantly outperforms the other approaches, including the two metric-based state-of-the-art methods for the 4-layer embedding:  Relation Nets\cite{RelationNet} and Prototypical Nets\cite{PrototypeNet}.

With the deeper 9-layer embedding (Table \ref{Tab:MiniImageNet}), we were able to obtain better results than the state-of-the art SNAIL method~\cite{Snail} with 13-layer embedding and all other methods to date (to the best of our knowledge). Given that SNAIL used a 13-layer residual network embedding which is considerably more powerful than our standard 9-layer CNN embedding, better performance may indicate that our method can utilize the CNN embedding much better than the SNAIL method.

\subsubsection{Ablation Analysis}
We conducted a set of ablation experiments to analyze the performance improvement due to class support embedding and due to the Approximate Ensemble of Meta-Learners. Results are reported in Table \ref{Tab:Ablation}.
In addition to the benchmark tasks, 5-way 1-shot and 5-way 5-shot classification, we also ran additional experiments on 5-way 2-shot classification tasks to gain more insights on how the proposed method behaves as the number of shots increases. It can be observed in Table \ref{Tab:Ablation}, that the gain in performance due to the class support embedding increases as the number of shots decreases. This result verifies that our proposed class support embedding is especially useful when fewer samples per class are present, in particular, in one-shot learning. This result supports our belief that when support samples are generated at random, and there are very few of them (one or two), they are unable to represent their class well. Our proposed class support network can learn to reposition/correct these random support points to better represent their class.

As can be seen in Table \ref{Tab:Ablation}, Approximate Ensemble of Meta-Learners consistently improved the results over the solution obtained by early stopping. AEML is especially beneficial for preventing overfitting in training large capacity models.

\section{Conclusion}
We proposed a simple method called a Class Support Network
for one/few-shot learning. Our method integrated a mechanism for learning better class representatives (given only the support set) with a dynamic prototype selection (given these new representatives and a query). Such a combination improved meter-learning part of the network and permitted a deeper CNN embedding. The network was trained end-to-end with episode training  and provided state-of-the-art results on one/few-shot benchmarks. Additionally, we proposed a very efficient approximation of ensemble learning which made episode training  easier and reduced overfitting.

For future work, we would like to investigate alternative prototype selection schemes. For example, probabilistic selection by regarding each prototype as a random sample of a component of a mixture of Gaussians. In doing so, all the support points would contribute to classification in a probabilistic manner, with potential further improvement in performance.








\bibliography{latent_support_net}
\bibliographystyle{ieeetr}

\section{Appendix}
\subsection{On Relation between AEML and model ensemble}\label{AppendA}
Given that top-$t$ points along the training trajectory are already good models and are close to the optimal point denoted as $x^{*}$, we can assume that the value of the function in each of the  top-$t$ models may be expressed as the value of the function in $x^{*}$ plus a linear perturbation, as follows
\begin{align}
	F(x) = F(x^{*}) + L(x, x^{*})
\end{align}
where $L$ is linear w.r.t $x$. One can easily see that
\begin{align}
	\frac{1}{k} \sum_{i =1}^{k} F(x_{i}) &= F(x^{*}) + \frac{1}{k} \sum_{i=1}^{k}  L(x_{i}, x^{*}) \\
    &= F(x^{*}) + L(\frac{1}{k} \sum_{i=1}^{k} x_{i}, x^{*}) \\
    &= F(\sum_{i=1}^{k} x_{i})
\end{align}
which means we can approximate ensemble of the top-$t$ models by averaging their weights.

\subsection{Experiments On Relation between AEML and model ensemble}\label{AppendB}
To validate the feasibility of approximating model ensemble with the proposed AEML, we conducted an additional experiment on \textit{miniImageNet} dataset. For the ensemble method, we chose a committee and allow the models to vote for the best prediction. We used the proposed class-support network with the standard 4-layer CNN embedding (as in \textit{Omniglot} and \textit{miniImageNet} experiments), but excluding the support set embedding, which is basically equivalent to Matching networks~\cite{MatchingNet} for one shot learning.  We tested it in 5-way 1-shot classification, and reported the accuracy improvement averaged over 2000 episodes. The accuracy improvement of ensemble method and the proposed AEML was 0.48\% and 0.39\% respectively which is quite similar.


\end{document}